%% file: main.tex
\pdfoutput=1
\documentclass[11pt]{article}

\usepackage[preprint]{acl}
\usepackage{graphicx}
\usepackage{multirow}
\usepackage{marvosym}
\usepackage{footmisc}
\usepackage{booktabs} % 用于更美观的表格线
\usepackage{times}
\usepackage{amsthm}
\usepackage{algorithm}
\usepackage{algpseudocode}
\usepackage{latexsym}
\usepackage{float} % 引入float宏包，用于禁止浮动体浮动
\usepackage{amsmath}
\usepackage[T1]{fontenc} % For proper rendering and hyphenation of words containing Latin characters (including in bib files)
\usepackage[utf8]{inputenc} % This assumes your files are encoded as UTF8
\usepackage{microtype}
\usepackage{inconsolata}
\usepackage{amsfonts}
\title{DictLLM: Harnessing Key-Value Data Structures with Large Language Models for Enhanced Medical Diagnostics}

\author{
  Yiqiu Guo$^{1,2}$,
  Yuchen Yang$^{2,4}$, 
  Ya Zhang$^{2,3}$\textsuperscript{\Letter},
  Yu Wang$^{2,3}$\textsuperscript{\Letter},
  Yanfeng Wang$^{2,3}$
  \\ [0.15cm]
  $^1$Fudan University~~~~$^2$Shanghai AI Laboratory~~~~$^3$Shanghai JiaoTong University\\
  $^4$University of Science and Technology of China \\
}

\begin{document}
\maketitle
\begin{abstract}
\footnotetext{\Letter: Corresponding author.}
Structured data offers a sophisticated mechanism for the organization of information. Existing methodologies for the text-serialization of structured data in the context of large language models fail to adequately address the heterogeneity inherent in key-value structured data. These methods are not ideal and frequently result in larger input sizes and poor adaptability to input changes. In this paper, we introduce DictLLM, an innovative framework designed to improve the modeling of key-value structured data, like medical laboratory reports, for generating medical diagnoses. DictLLM integrates three key components: (1) group positional encoding to maintain permutation invariance, (2) hierarchical attention bias to capture the inherent bias in structured data, and (3) an optimal transport alignment layer that aligns the embedding generated by the dictionary encoder with the LLM, thereby producing a sequence of fixed-length virtual tokens. We carry out experiments using various LLM models on a comprehensive real-world medical laboratory report dataset for automatic diagnosis generation, our findings illustrate that DictLLM significantly outperforms established baseline methods and few-shot GPT-4 implementations in terms of both Rouge-L and Knowledge F1 scores. Furthermore, our evaluation of the framework's scalability and robustness, through a series of experiments, underscores its exceptional capability in accurately modeling the complex key-value data structure of medical dictionary data.
\end{abstract}

\input{sections/intro_v2}
\input{sections/related_work}
\input{sections/approach}
\input{sections/experiment_setup}
\input{sections/results}
\input{sections/conclusion}

\bibliography{custom}

\appendix
\input{sections/appendix.tex}

\end{document}

%% file: sections/intro_v2.tex
\section{Introduction}

\begin{figure}[t!]
  \centering
  \includegraphics[width=0.5\textwidth]{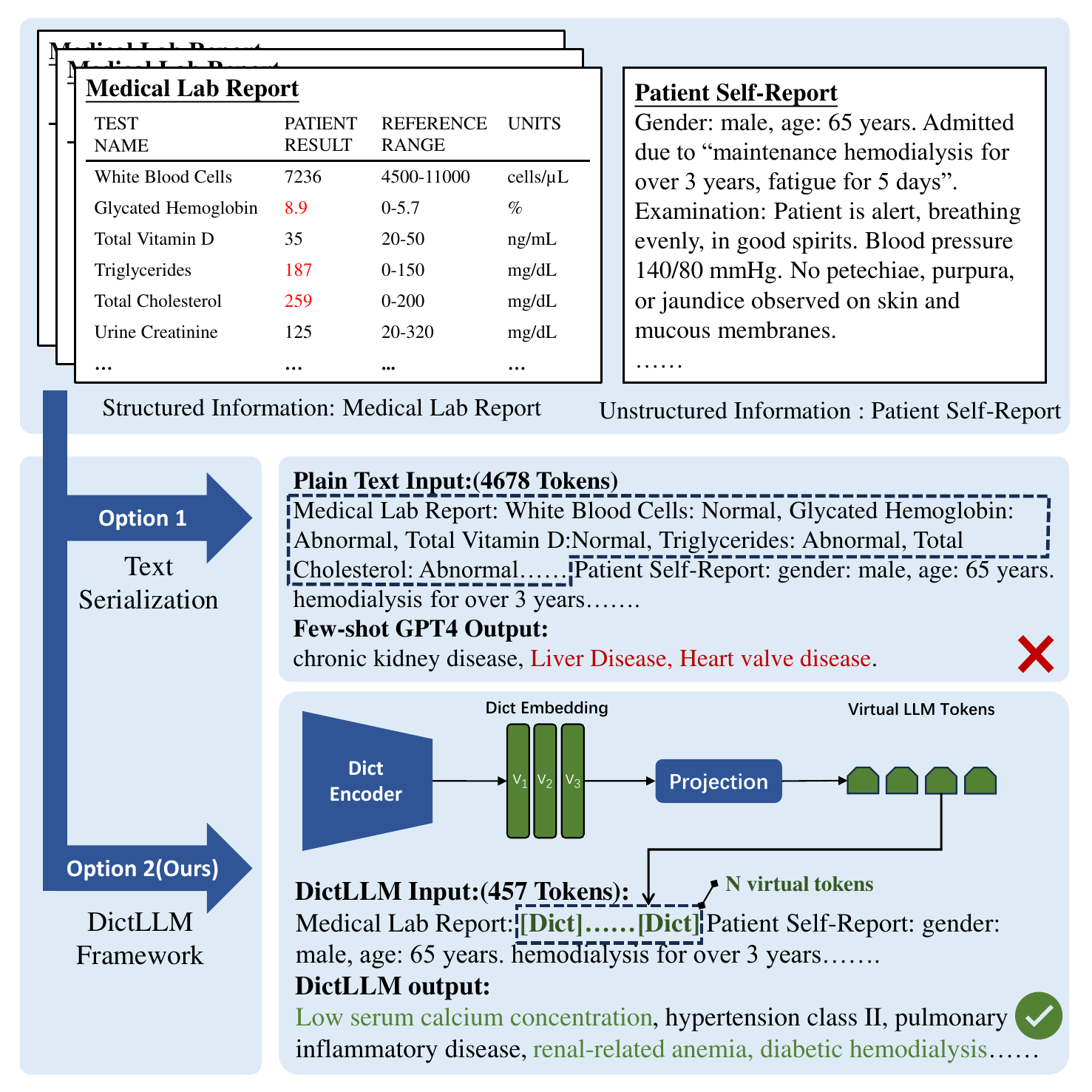}
  \caption{Our DictLLM Framework for medical lab report-assisted diagnosis generation. 
  The framework uses a hierarchical dict encoder to encode the medical lab report, and an optimal transport alignment layer to align the embedding generated by the dict encoder and the text encoder.}
  \label{fig:fig_intro}
\end{figure}

The integration of large language models (LLMs) into natural language processing (NLP) has marked a paradigm shift, enabling unprecedented advancements across diverse applications. Recent explorations into applying LLMs to structured data processing, such as graphs, dictionaries, and tables, highlight their potential beyond traditional text analysis. Notably, efforts like  tabular data classification in \citet{TabLLMFewshot-2023}, graph-based node classification in \citet{GraphGPTGraph-2023} , and intelligent Excel table querying in \citet{KuBiaoChatExcel-2023} , have paved the way for innovative applications. Yet, the application of LLMs in processing medical lab reports, a cornerstone in clinical diagnostics, exposes significant challenges. These reports, structured as key-value pairs, are critical for diagnosis but diverge substantially from the data types traditionally handled by LLMs due to their unique structure and information content.

Medical lab reports are pivotal in clinical decision-making, capturing patient test results in a structured format that facilitates diagnosis. Unlike the linear, narrative flow of natural language, these reports are characterized by two distinct features:
\noindent
\begin{itemize}

\item Structural Heterogeneity: They are organized as key-value pairs, allowing for permutation invariance where the sequence of entries does not affect the informational content.
\item Information Density Heterogeneity: These reports encapsulate densely packed, discrete data, contrasting with the more continuous and narrative nature of text.

\end{itemize}

Existing methods, primarily based on converting structured data into a linear token sequence, inadequately capture these nuances. Such serialization not only risks losing structural fidelity but also scales poorly due to token limits in LLMs, highlighting a critical gap in current methodologies.

DictLLM emerges as a novel framework tailored to address these challenges, marrying the structured precision of medical lab reports with the analytical depth of LLMs. By innovatively leveraging a hierarchical dict encoder inspired by advancements in set transformation, DictLLM transcends traditional serialization approaches. It introduces a dict tokenizer to convert complex numerical data into interpretable medical labels, a group positional encoding to maintain the inherent permutation invariance of lab report data, and hierarchical attention mechanisms to adeptly handle the reports' information density.

Our contributions offer a significant leap forward in medical diagnostics:
\noindent
\begin{itemize}
\item We introduce a hierarchical dict encoder that adeptly models the structured nature of medical lab reports, preserving their key-value integrity and enhancing robustness to variations in report formatting.

\item The introduction of an optimal transport alignment layer aligns dict encoder embeddings with LLM outputs, optimizing the efficiency of input representation and addressing the challenge of token count scalability.

\item Comparative analysis with leading LLMs on a comprehensive dataset of real-world medical lab reports demonstrates DictLLM's superior performance, showcasing notable improvements in Rouge-L and Knowledge F1 scores, indicative of its enhanced diagnostic accuracy and relevance extraction capabilities.

\end{itemize}

In aligning closely with the medical diagnostic process's intricacies, DictLLM not only highlights the untapped potential of LLMs in processing structured medical data but also sets a new benchmark for precision and efficiency in automated medical diagnosis. This approach not only underscores the framework's novelty but also its practical significance, promising to bridge the gap between current LLM capabilities and the complex demands of healthcare diagnostics.

%% file: sections/related_work.tex
\section{Related work}
\subsection{Tabular data representation learning}

Tabular data representation learning aims to learn a dense representation for tabular data.  \citet{TURLTable-2020} introduces the Masked Entity Recovery (MER) objective for pre-training the Table Encoder, aiming to capture the semantics and knowledge in large-scale unlabeled data. \citet{TableFormerRobust-2022} highlights that linearizing table structures would encode the order of the table's rows and columns with an unwanted bias. \citet{HYTRELHypergraphenhanced-2023} introduces a hypergraph-enhanced table representation learning framework to model the inherent inductive bias of tabular structures. \citet{CTBERTLearning-2023} introduce cross-table pretraining into the tabular data representation learning, to capture the cross-table knowledge. \citet{LearningEnhanced-2022} propose learning enhanced representations for tabular data via neighborhood propagation. These study highlights the importance of modeling the structural properties of tabular data. However, these approaches do not harness the capabilities of large language models and are not designed to explicitly capture the heterogeneity of medical lab reports.

\subsection{Large language model for structural data}

With the emergence of large language models \citet{LLaMAOpen-2023} \citet{GLM130BOpen-2022} \citet{TrainableOptimal-2021}, there have been numerous efforts to leverage them for processing structured data tasks. \citet{ChartLlamaMultimodal-2023} propose ChartLlama, a multimodal llava-based model for chart understanding and generation task. \citet{TabLLMFewshot-2023} introduce TabLLM, an text serialization-based framework that leverages LLMs for data-efficient tabular classification. However, this approach can only handle small-scale classification tasks, which is not suitable for generation tasks. \citet{OpenTabAdvancing-2023}propose OpenTab, an open-domain end-to-end table reasoning framework, which leverages a retriever to fetch relevant tables, employs a coder to generate programs as intermediary reasoning steps, and assigns the task of deriving the final solution to a reader. However, the retrieval-augmented paradigm can be limited by the performance of the retrieval module, especially for tasks requiring specific domain knowledge. \citet{TATLLMSpecialized-2024} propose TAT-LLM, A specialized language model for discrete reasoning over tabular and textual data, serve as a pioneering example of specializing smaller language models for specific tasks. The GraphGPT proposed by \citet{GraphGPTGraph-2023} comes closest to our work. This method employs a graph encoder and a text encoder to encode the structural information and the textual information of the graph and propose a dual-stage graph instruction tuning paradigm. Our work distinguishes itself from these studies by focusing on the design of a carefully designed hierarchical dict encoder to model the heterogeneous structure of medical lab reports.

%% file: sections/approach.tex
\section{Approach}

\subsection{Problem Formalization}

\begin{figure}[h]
  \centering
  \includegraphics[width=0.49\textwidth]{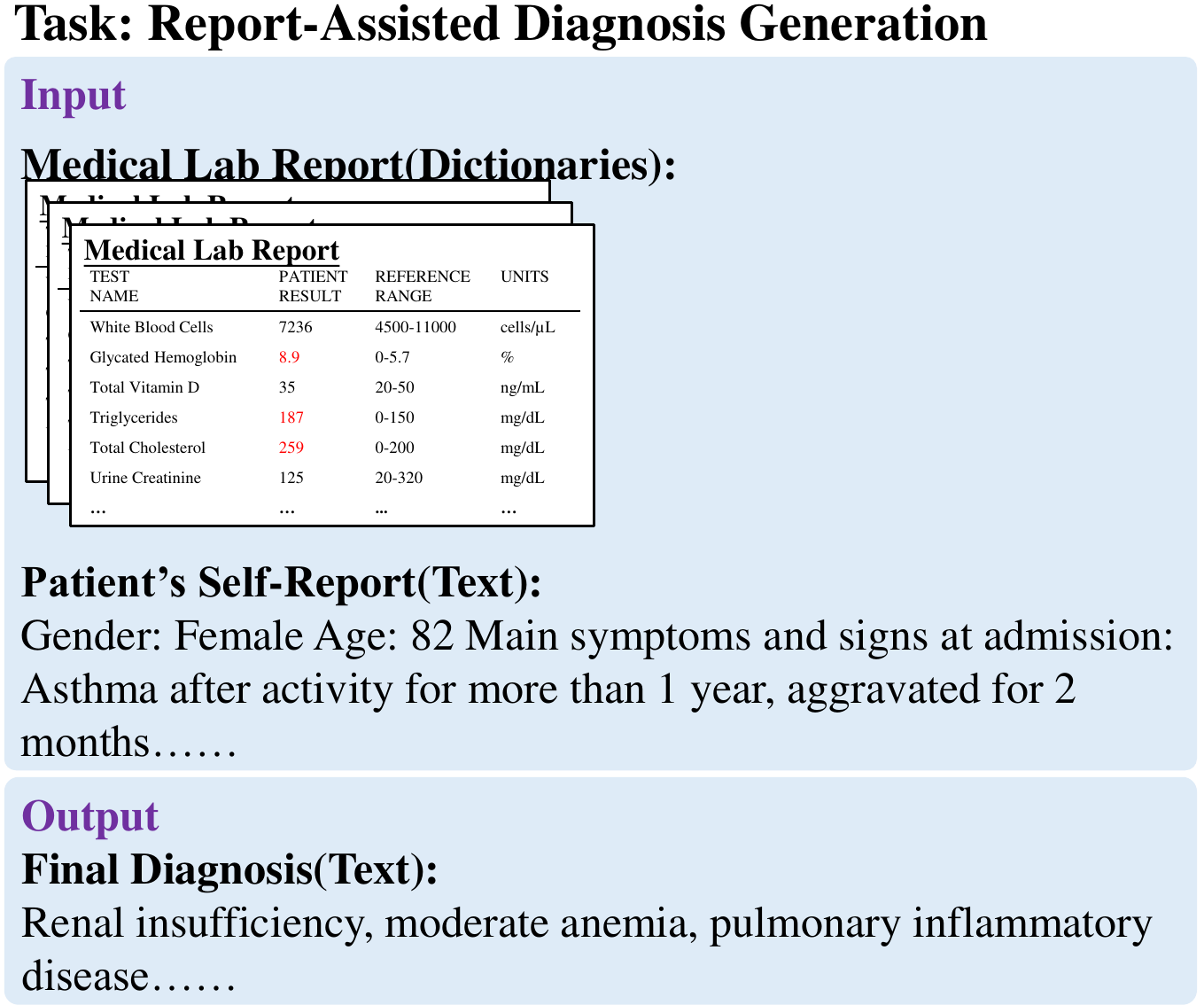}
  \caption{
    An example of the input and output of the medical lab report-assisted diagnosis generation task.
  }
  \label{fig:fig_io}
\end{figure}

As shown in Figure \ref{fig:fig_io}, the task of report-assisted diagnosis generation involves creating a diagnosis based on a patient's self-reported symptoms and medical laboratory reports. Suppose we have a patient's medical laboratory report. We can formalize this report as a set of dictionaries, denoted as $\mathcal{D} = \{D_1, D_2, \ldots, D_n\}$, where each $D_i$ can be formalized as $D_i = \{(k_{ij}, v_{ij})\}^m_{j=1}$, where $k_{ij}$ and $v_{ij}$ represent the key and value of the $j$-th key-value pair in the $i$-th dictionary, respectively. The text information of the patient's self report can be formalized as a sequence of tokens, denoted as $\mathcal{T} = \{t_1, t_2, \ldots, t_m\}$, where $t_i$ represents the $i$-th token in the sequence. The goal of the report-assisted diagnosis generation task is to generate the final diagnosis of the patient, denoted as $\mathcal{Y} = \{y_1, y_2, \ldots, y_k\}$, where $y_i$ represents the $i$-th token in the sequence.

\begin{figure*}[h!]
  \centering
  \includegraphics[width=\textwidth]{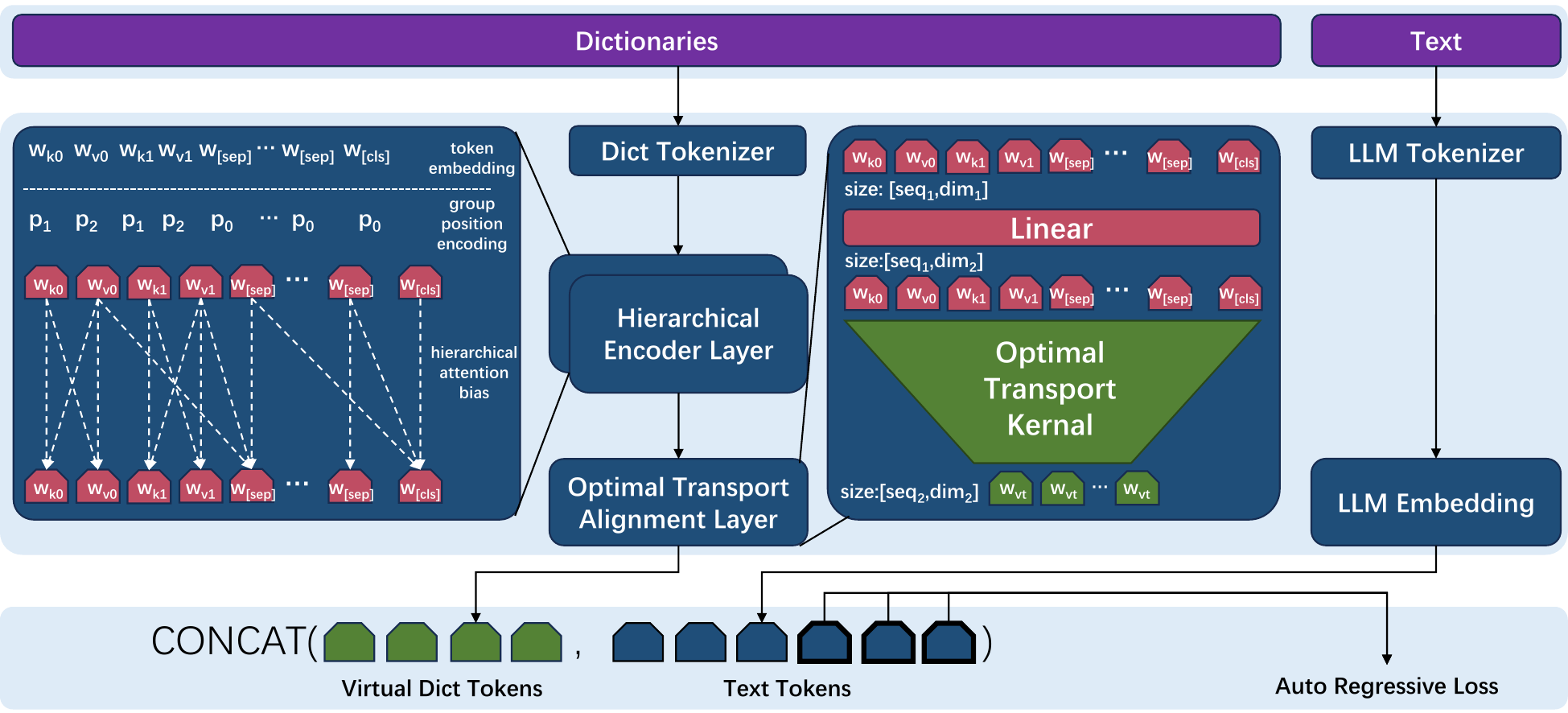}
  \caption{DictLLM Framework for report-assisted diagnosis generation. The medical lab report is first tokenized and encoded by the dict encoder. The embedding generated by the dict encoder are then aligned with the text embedding generated by the large language using the optimal transport alignment layer. The aligned embedding are then fed into the large language model to generate the final diagnosis.}
  \label{fig:fig_framework}
\end{figure*}

\subsection{Framework}

In the pipeline of a text-serialization based method, the dictionaries are converted into a single natural-language string using a fixed template. However, this approach is sub-optimal for structured data like dictionary due to the structural heterogeneity between structured data and natural language.

To address this, We propose the DictLLM Framework. As shown in Figure \ref{fig:fig_framework}, the DictLLM Framework consists of three main components: a hierarchical dict encoder, an optimal transport alignment layer, and a large language model. The hierarchical dict encoder and optimal transport alignment layer encode medical laboratory reports into several virtual tokens $\mathcal{T}_v$, the virtual tokens are then concat with the text tokens $\mathcal{T}$, and the combined tokens are fed into a large language model for generation.

\subsection{Hierarchical Dict Encoder}

Drawing inspiration from recent advancements such as SetTransformer \citet{SetTransformer-2019}, TURL \citet{TURLTable-2020}, and Tapas \citet{TaPasWeakly-2020}, we harness the BERT’s self-attention architecture  \citet{BertPretraining-2018} to model the intricate interactions within dictionaries. To effectively adapt to the unique data attributes of medical laboratory reports, the dict encoder incorporates dict tokenizer, relative position encoding and hierarchical attention biases. In the following sections, we will describe the them in detail.

\subsubsection{Dict Tokenizer: tokenize numerical values in lab report}
Dict tokenizer turns dictionaries into a series of token ids. To align with the behavior of medical practitioners in actual medical practice, we propose converting detailed numerical values in the laboratory reports into special medical labels. For a numerical attribute \(v_{ij}\), the dict tokenizer maps it to a single token \(v'_{ij}\) (e.g., [NORMAL], [POSITIVE], [NEGATIVE]). We have defined a total of 13 such special medical labels, with a detailed list provided in the appendix. To be more specific, given a set of dictionaries $\mathcal{D} = \{D_1, D_2, \ldots, D_k\}$ that contains $k$ dictionary, where each $D_i = \{(k_{ij}, v_{ij})\}^m_{j=1}$ is a set that contains $m$ key-value pairs. The dict tokenizer function $f_{t}$ maps the whole $\mathcal{D}$ into a series of token ids $\textit{t}$, denoted as:
$$f_{t}(\mathcal{D}) \rightarrow \textit{t} = \{t_1, t_2, \ldots, t_n\}$$

\subsubsection{Group Positional Encoding: maintain permutation invariance}
After tokenization, the discrete token ids will be embedded into continuous vectors, which will be fed into the hierarchical encoder layer. We follow the standard practice of using a token embedding $W$ and add a positional encoding $P$ to the token embedding.

To model the permutation invariance of key-value pairs in laboratory reports, we have established a group positional encoding $P_{group}=\{p_{\text{pos}_1}, p_{\text{pos}_2}, \dots, p_{\text{pos}_n}\}$. This encoding ensures that perturbation in the relative positions of elements within a dictionary do not impact the embedding generated by the dict encoder. Given the distinct characteristics of medical laboratory reports as dictionary-structured data, We propose the following assumption:

\noindent
\textbf{Assumption}: \textit{
  For a laboratory report \(D\) containing \(m\) key-value (\(k, v\)) pairs, changing the relative positions of these (\(k, v\)) pairs within \(D\) does not affect the final diagnosis.
}

We implement $P_{group}$ by resetting  the index of positional ids at the beginning of each key-value pair, where $\text{pos}_i$ represents the positional id for the \(i\)th token. Let $W_{emb}$ be the embedding matrix of the dict encoder, The initial dict embedding $h_0$ is denoted as:
$$h_0=W_{emb}(\textit{t})+P_{group}$$

\subsubsection{Hierarchical Attention Bias: model structural inductive bias}
Medical laboratory reports distinguish themselves from natural language in that, the correlation among items within a single report is significantly stronger than the correlation among items across different reports. (e.g. Test items on the same urine report are more likely to collectively indicate kidney-related diseases) We propose incorporating hierarchical attention bias to model the structural inductive bias of medical laboratory reports.

Specifically, tokens within the same dictionary are visible to each other, while tokens from different dictionaries are not. The special token \text{[sep]} is used to separate different dictionaries, and the special token \text{[cls]} is used to represent the whole dictionary. These special tokens are visible to each other, and they are visible to all tokens in their own dictionary. As illustrated in the Figure \ref{fig:fig_framework}, tokens connected by dashed lines are visible to each other, while others are not.

The initial embedding will then pass through multiple hierarchical encoder layers(HierEnc) to obtain the final embedding. The hierarchical attention bias is implemented as a attention mask $M$, which is a $n \times n$ matrix, where $n$ is the sequence length. A hierarchical encoder layer consists of a Hierarchical Self-Attention (HierAttn) layer and a MLP layer, denoted as:
\begin{align*}
  \small
  \textit{HierEnc}(h_l)                        = h_l + \textit{HierAttn}(h_l) + \textit{MLP}(h_l) \\
  \textit{HierAttn}(h_l)              = \textit{softmax}(\frac{QK^T+M}{\sqrt{d_K}})V              \\
  M_{ij}                         = \begin{cases} 1 & t_i, t_j \in D \\ 1 & t_i, t_j \in \{\textit{[sep]},\textit{[cls]}\} \\ 0 & otherwise\end{cases}
\end{align*}

After passing through the hierarchical encoder layers, the final dict embedding $h_{L}$ is obtained.

\subsection{Optimal Transport Alignment Layer}

\begin{algorithm}[h]
  \caption{Optimal Transport Alignment Layer}
  \label{alg:algorithm1}
  \begin{algorithmic}[1]
    \Statex \hspace*{-\algorithmicindent}\textbf{Input} source embedding $h_s \in \mathbb{R}^{m\times a}$
    \Statex \hspace*{-\algorithmicindent}\textbf{Output} target embedding $h_t \in \mathbb{R}^{n\times b}$
    \State initialize (trainable) reference points $z \in \mathbb{R}^{n\times b}$
    \State initialize positive definite kernel $\varPhi$.
    \State $h_r \in \mathbb{R}^{m\times b} \longleftarrow \varPhi(h_s)$
    \State $\textit{TP} \in \mathbb{R}^{n\times m} \longleftarrow \textit{sinkhorn}(h_r, z)$
    \State $h_t \in \mathbb{R}^{n\times b} \longleftarrow \textit{TP} \times h_r $
  \end{algorithmic}
\end{algorithm}

To deal with the heterogeneous information density between medical laboratory reports and natural language, we propose an optimal transport alignment layer to align the embedding generated by the dict encoder with those generated by the LLM, producing a list of fixed-length virtual tokens.

Natural language organized information in a sequential, dense and coherent manner, while information in medical laboratory reports are sparse and discrete. A naive approach such as using a linear layer may not be the optimal solution. Optimal transport is a mathematical framework that provides a principled way to align two sets of points in a high-dimensional space, which is widely used to alignment problems. \citet{UnsupervisedAlignment-2018}

We utilize a recently proposed technique called optimal transport kernel \citet{TrainableOptimal-2021} (OTK). OTK first first utilize a positive definite kernel (i.e. in our implementation, a linear function) to embed the source set into a reproducing kernel Hilbert space (RKHS), then sinkhorn algorithm, which is a differentiable approximation of the optimal transport plan, is used to compute the optimal transport plan between the source set and a trainable reference set, which introduce non-linear transformation on source features. The detailed process of is described in algorithm \ref{alg:algorithm1}.

Let the embedding output by the dict encoder be denoted as $h_L \in \mathbb{R}^{m\times a}$, where $a$ is the number of tokens in the dict embedding, and $b$ is the dimension of the token embedding. Our goal is to map it to a fixed-length virtual token $\mathcal{T}_v = \{t_{v1}, t_{v2}, \ldots, t_{vn}\} \in \mathbb{R}^{n\times b}$, where $n$ is the number of virtual tokens, and $b$ is the dimension of the large language model's token embedding.

%% file: sections/experiment_setup.tex
\section{Experiments Setup}

\subsection{Data Description}
\begin{figure}[H]
    \centering
    \includegraphics[width=0.5\textwidth]{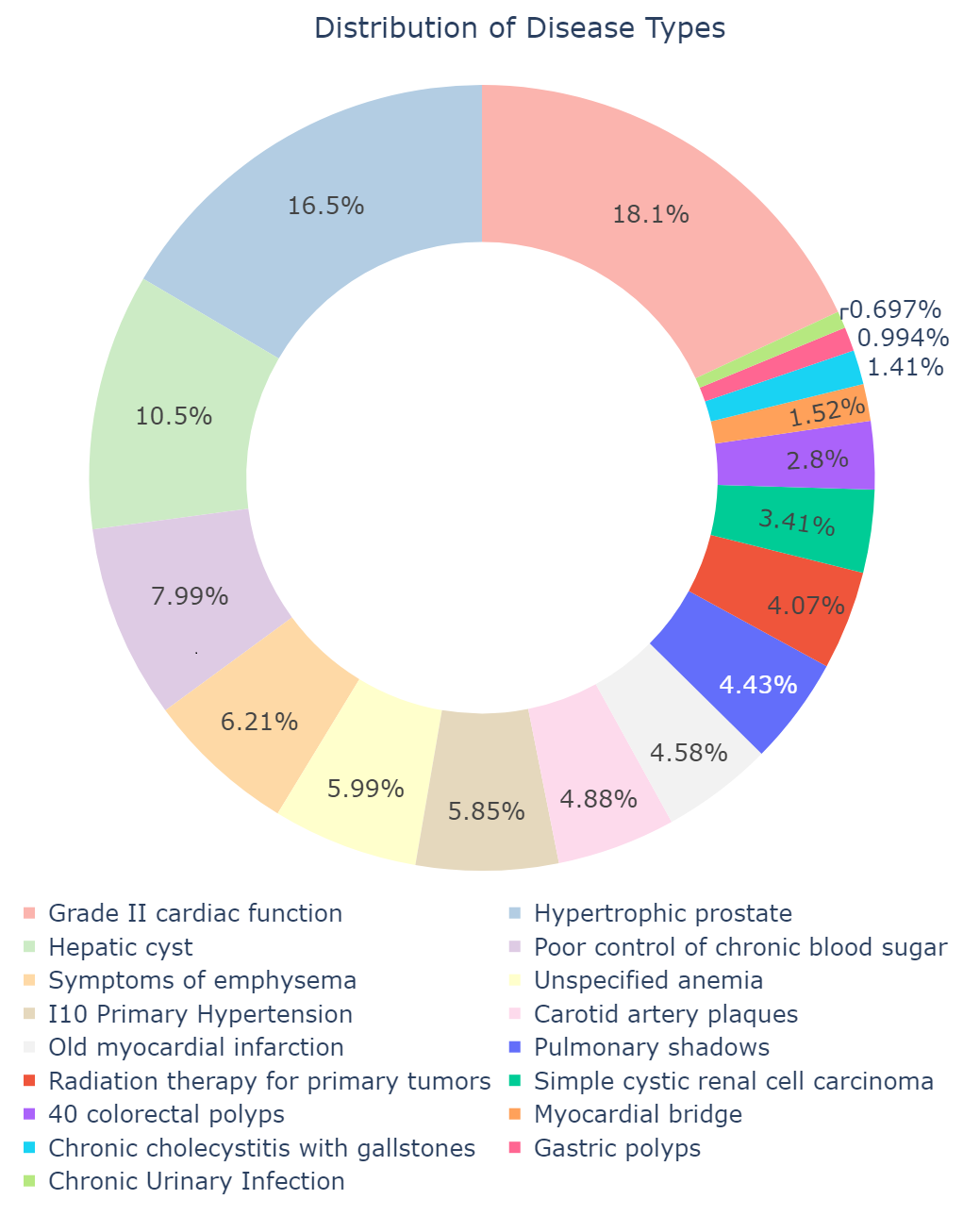}
    \caption{
        Distribution of different types of disease in the dataset.
    }
    \label{fig:fig_disease_type}
\end{figure}
The dataset we use in our experiment is a large-scale chinese real-world medical lab report dataset. We collect the dataset from a real-world hospital, which contains a large number of medical lab reports and the corresponding final diagnosis. The dataset contains a total of 11, 290 medical lab reports, and each report is associated with several final diagnosis. The original dataset is highly imbalanced in terms of the number of the disease types. We only keep the disease types that appears more than 0.1\% of the time in the dataset. The dataset contains a wide range of disease types, as shown in Figure \ref{fig:fig_disease_type}. The statistics of the dataset are shown in Table \ref{tab:dataset_description}.

\input{tables/data_description.tex}

\subsection{Baseline Methods}

\textbf{Text-Serialization}
For text-serialization method, we use a fixed template to serialize the medical lab report into a sequence of tokens. In our experiment, we separate each item in dict with comma, and use a special token to separate each dict. Then the model will be trained with the standard supervised fine-tuning paradigm.

\noindent
\textbf{GPT-4}
We also evaluate the performance of GPT-4 on this task in zero-shot and few-shot settings. The template we use is the same as the one we use in baseline.

\subsection{Implementation Details}
For our model implementation, we primarily rely on the PyTorch and Transformers libraries. In terms of the Text-Serialization method, we convert medical lab reports into plain text at the dataset level and then train the model using the standard supervised fine-tuning paradigm. For our proposed DictLLM framework, we train the dict encoder and base large language models jointly. We choose internlm-7b-base and baichuan2-7b-base as our base models due to their superior performance in Chinese. We utilize the AdamW optimizer with a learning rate of \(2e-5\) and a total batch size of 128. We apply a warmup ratio of 0.01, and the training process spans 6 epochs. Notably, we did not conduct any hyperparameter search in our experiment. Regarding the dataset, we split it into training and testing sets, using 90\% of the data for training and the remaining 10\% for testing.

\subsection{Evaluation Metrics}
\label{subsec:metrics}
We use the following metrics to evaluate the performance of the methods we proposed in this paper:

\noindent
\textbf{Rouge-L}
Rouge-L is a metric that measures the similarity between two sequences. It is widely used in the text generation task.

\noindent
\textbf{Knowledge F1}
We also use the knowledge F1 score to evaluate the performance of the methods we proposed in this paper. Knowledge F1 score is a metric that measures the quality of the generated sequence in terms of the knowledge it contains. In our experiment, we implement the knowledge F1 score as the harmonic mean of precision and recall of the correct diagnosis in the generated sequence.

%% file: tables/data_description.tex
\begin{table}[H]
    \centering
    \resizebox{0.45\textwidth}{!}{
        \begin{tabular}{ccc}
            \toprule
            num of  & mean num of     & mean num of          \\
            cases   & text token/case & lab report item/case \\
            \midrule
            11,290  & 450.82          & 16.23                \\
            \bottomrule
        \end{tabular}
    }
    \caption{
        Statistical information of datasets.
    }
    \label{tab:dataset_description}
\end{table}

%% file: sections/results.tex
\section{Results}

\subsection{Main Results}

\input{tables/main_results_v2.tex}

Table \ref{tab:main_results} shows the main results of our experiment. As we can see, the proposed DictLLM framework outperforms the baseline methods in terms of both Rouge-L and Knowledge F1 score in all settings. The performance of our method is consistent across different backbone models. The results demonstrate that our proposed DictLLM framework is effective in modeling the heterogeneous structure of medical lab reports and generating the final diagnosis.

Notably, GPT-4 achieve poor performance in both zero-shot and few-shot settings, and there is a large gap between the performance of GPT-4 and the finetuned large language models. The main reason is report-assisted diagnosis generation task is that the task requires the model to have a good understanding of the specialized medical terminology, which is rare in the training data of GPT-4. 

The gap between the performance of the text-serialization method and our proposed DictLLM framework in baichuan-7b is smaller than that in internlm-7b, which is mainly due to the better backbone model performance of baichuan-7b. 

\subsection{Scalability to Input Length}

\begin{figure}[h]
    \centering
    \includegraphics[width={0.5\textwidth}]{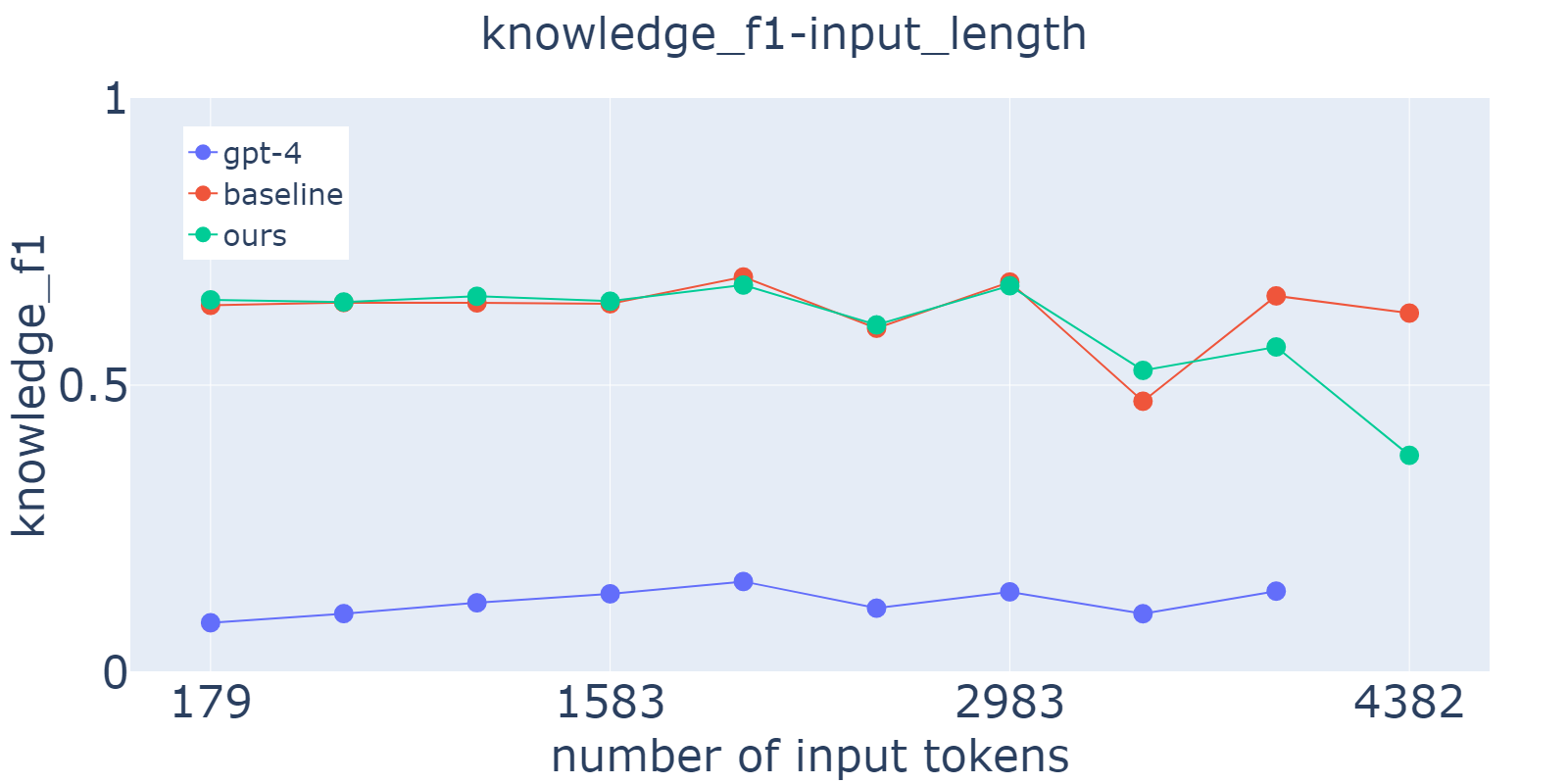}
    \caption{
        The knowledge F1 score of different methods with respect to the number of input tokens. Other results are detailed in the appendix \ref{sec:appendix}.
    }
    \label{fig:fig_knowledge_f1}
\end{figure}

We also evaluate the scalability of several method to the input length on the backbone of internlm-7b. In real-world medical lab reports, the number of items in the report can be very large, and the length of the report may exceed the limitation of the max token length of large language models. Large input length would also lead to large training time and memory requirement, which could be a bottleneck for the model to be deployed in real-world applications.

As shown in Figure \ref{fig:fig_knowledge_f1}, the performance of the text-serialization method decreases significantly as the input length increases due to the large input token size. In contrast, our proposed DictLLM framework effective compress the input token number and achieve consistent performance across different input lengths, demonstrating a better scalability of our method to the input length.

\subsection{Robustness to Input Perturbation}
\input{tables/perturbation_results.tex}
Besides scalability, Robustness to input perturbation is also an important property for the model to be deployed in real-world applications. Input perturbation refers to the random permutation of the items in the medical lab report. In the ideal situation, the model should generate the same diagnosis for the same medical lab report, regardless of the order of the items in the report. To evaluate the robustness of the model to input perturbation, we conduct an experiment to compare the performance of different methods before and after perturbation. We report the performance and the relative change of the generated text before and after perturbation in Table \ref{tab:perturbation_results}. The metric RC (i.e. Relative Change) is calculated as the $1 - RougeL_{f1}$ score between the text generated before and after the perturbation.

As is shown in Table \ref{tab:perturbation_results}, the performance of the text-serialization method decreases after perturbation, while the performance of our proposed DictLLM framework is the most stable across different backbone models. We also observe that the relative change of the generated text before and after perturbation is the smallest for our proposed DictLLM framework, demonstrating the robustness of our method to input perturbation.

\subsection{Ablation study}

\subsubsection{Ablation over the main components}

\input{tables/ablation_results.tex}

We conduct ablation study to demonstrate the effectiveness of the model components in our proposed DictLLM framework. For the ablation of group positional encoding, we replace it with the standard sequential positional encoding. For the ablation of optimal transport alignment layer, we replace it with the a simple linear layer. For the ablation of hierarchical attention bias, we just simply remove it from the model. Table \ref{tab:ablation_results} shows the ablation study results.

Overall, the results show that each component in our proposed DictLLM framework contributes to the performance of the model. Among all the components, deleting the hierarchical attention bias leads to the largest performance drop, demonstrating the importance of the hierarchical attention bias in capturing the structural inductive bias of medical lab reports.

\subsubsection{Ablation over the virtual token number}

\begin{figure}[h]
    \centering
    \includegraphics[width={0.5\textwidth}]{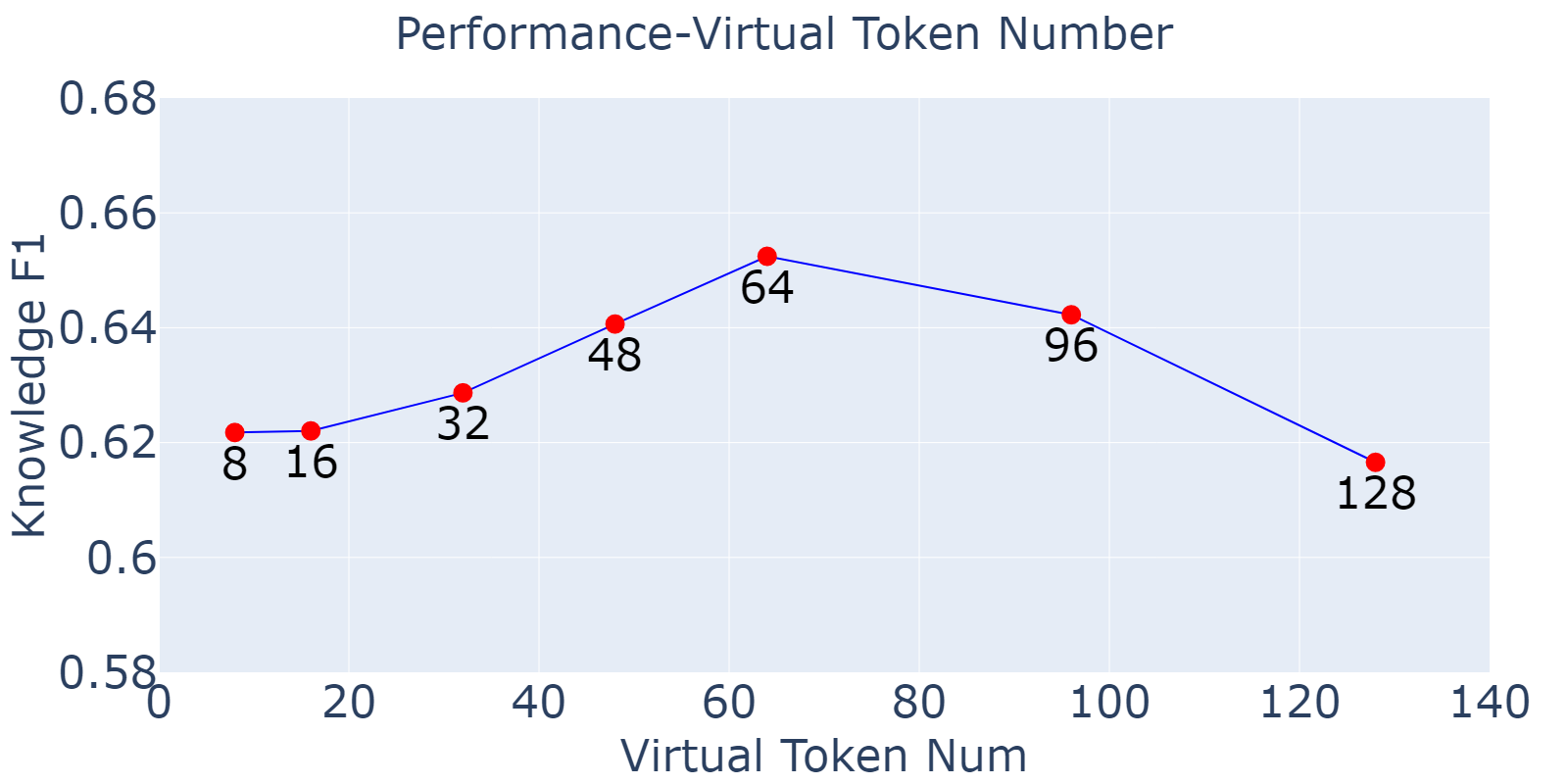}
    \caption{
        Ablation study of virtual token ber.
    }
    \label{fig:fig_knowledge_f1}
\end{figure}

The number of the virtual token is a hyperparameter in our proposed DictLLM framework. We conduct an ablation study to evaluate the performance of the model with different virtual token number. As shown in Figure \ref{fig:fig_knowledge_f1}, the performance of the model increases as the virtual token number increases.

However, the increase of the virtual token number also leads to the slightly increase of the model size and the memory requirement. We choose 64 as the virtual token number in our experiment, which achieves a good trade-off between the performance and the memory requirement.

%% file: tables/main_results_v2.tex
\begin{table*}[h]
    \centering
    \tiny
    \resizebox{0.95\textwidth}{!}{
        \begin{tabular}{ll|cccccc}

            \toprule
            \multicolumn{2}{c}{\multirow{2}{*}{Method}} & \multicolumn{3}{c}{Rouge-L} & \multicolumn{3}{c}{Knowledge}                                                                             \\
            \multicolumn{2}{c}{}                        & P                           & R                             & F1             & P              & R     & F1                              \\
            \midrule
            \multirow{2}{*}{\textsc{gpt-4}}             & zero-shot                   & 5.65                          & 32.87          & 8.64           & 8.82  & 28.85          & 12.70          \\
                                                        & few-shot                    & 5.80                          & 33.45          & 8.99           & 9.54  & 32.11          & 13.84          \\
            \midrule
            \multirow{4}{*}{\textsc{internlm-7b}}       & zero-shot                   & 3.59                          & 4.25           & 3.83           & 4.38  & 3.95           & 4.05           \\
                                                        & few-shot                    & 5.89                          & 5.35           & 5.29           & 5.99  & 5.32           & 5.53           \\
                                                        & finetune                    & 51.89                         & 45.19          & 46.69          & 50.90 & 46.03          & 47.43          \\
                                                        & DictLLM                     & 68.67                         & 63.15 & \textbf{64.24} & 68.68 & 64.09 & \textbf{65.24}                 \\
            \midrule
            \multirow{4}{*}{\textsc{baichuan-7b}}       & zero-shot                   & 6.14                          & 8.24           & 6.93           & 7.71  & 7.39           & 7.40           \\
                                                        & few-shot                    & 8.35                          & 12.67          & 9.83           & 10.19 & 9.27           & 9.58           \\
                                                        & finetune                    & 67.15                         & 63.18          & 63.13          & 67.32 & 64.42          & 64.51          \\
                                                        & DictLLM                     & 67.26                         & 63.39          & \textbf{63.28} & 67.50 & 64.65          & \textbf{64.61} \\
            \bottomrule
        \end{tabular}
    }
    \caption{
        \textbf{Main Results} We compare the performance of our DictLLM framework with several baseline methods on the medical lab report-assisted diagnosis generation task. We report the Rouge-L and Knowledge F1 scores. The best results are in bold. The detail of the evaluation metrics can be found in section\ref{subsec:metrics}.
    }
    \label{tab:main_results}
\end{table*}

%% file: tables/perturbation_results.tex
\begin{table*}[h]
    \centering
    \small
    \resizebox{\textwidth}{!}{
        \begin{tabular}{lcccccc|ccccccc}
            \toprule
            \multirow{3}{*}{Method}       & \multicolumn{6}{c}{Before Perturbation} & \multicolumn{7}{c}{After Perturbation}                                                                                                                                                        \\
                                          & \multicolumn{3}{c}{Rouge-L}             & \multicolumn{3}{c}{Knowledge}          & \multicolumn{3}{c}{Rouge-L} & \multicolumn{3}{c}{Knowledge} & RC$\downarrow$                                                                         \\
                                          & P                                       & R                                      & F1                          & P                             & R              & F1    & P     & R     & F1    & P     & R     & F1    &               \\
            \midrule
            \textsc{GPT-4}                & 5.65                                    & 32.87                                  & 8.64                        & 8.82                          & 28.85          & 12.70 & 5.77  & 32.21 & 8.64  & 8.79  & 27.69 & 12.43 & 36.30         \\
            \midrule
            $\textsc{Text-Serialization}$ & 53.43                                   & 46.31                                  & 47.76                       & 51.89                         & 47.43          & 48.60 & 52.69 & 45.77 & 47.24 & 52.22 & 47.63 & 48.82 & 11.31         \\
            $\textsc{DictLLM}$            & 68.53                                   & 63.52                                  & 64.22                       & 68.42                         & 64.40          & 65.11 & 68.61 & 63.64 & 64.33 & 68.51 & 64.49 & 65.20 & 1.71 \\
            % \midrule
            % $\textsc{Text-Serialization}_{\text{baichuan}}$ & 68.65                                   & 65.17                                  & 64.95                       & 68.70                         & 66.31          & 66.07 & 68.47 & 65.05 & 64.78 & 68.66 & 66.23 & 66.00 & 8.72          \\
            % $\textsc{DictLLM}_{\text{baichuan}}$            &                                         &                                        &                             &                               &                &       &       &       &       &       &       &       &               \\
            \bottomrule
        \end{tabular}
    }
    \caption{
        \textbf{Perturbation Results} We compare the performance of our DictLLM framework with baseline methods on the medical lab report-assisted diagnosis generation task before and after perturbation. RC denotes the relative change of the generated text before and after perturbation.
    }
    \label{tab:perturbation_results}
\end{table*}

%% file: tables/ablation_results.tex
\begin{table}[h]
    \centering
    \resizebox{0.5\textwidth}{!}{
        \begin{tabular}{l|cccccc}
            \toprule
            \multirow{2}{*}{Method} & \multicolumn{3}{c}{Rouge-L} & \multicolumn{3}{c}{Knowledge}                                                   \\
                                    & P                           & R                             & F1             & P     & R     & F1             \\
            \midrule
            \textsc{DictLLM}        & 68.67                       & 63.15                         & 64.24 & 68.68 & 64.09 & 65.24 \\
            \midrule
            - position encoding     & 67.25                       & 60.96                         & 62.23          & 67.29 & 62.23 & 63.46          \\
            - attention bias        & 66.15                       & 60.61                         & 61.40          & 66.19 & 61.69 & 62.53          \\
            - alignment layer       & 69.09                       & 61.36                         & 63.38          & 69.13 & 62.45 & 64.55          \\
            \bottomrule
        \end{tabular}
    }
    \caption{
        Ablation study of DictLLM framework.
    }
    \label{tab:ablation_results}
\end{table}

%% file: sections/conclusion.tex
\section{Conclusion}
In this paper, We propose a novel framework called DictLLM, which is an efficient and effective framework for modeling the heterogeneous structure of structured data, to deal with the report-assisted diagnosis generation task. Our comprehensive empirical studies on real-world datasets reveal that a carefully designed encoder, which individually encodes structured data, significantly enhances model performance on downstream tasks, demonstrating advantages in scalability and robustness.

\noindent
\textbf{Limitation} 
The DictLLM framework is specifically designed for processing dictionary-structured data and requires some effort to further extend it to more complex tabular data. Additionally, although DictLLM has reduced training and inference overhead compared to text-serialization methods, it still demands significant computational resources.

\section*{Acknowledgements}
This research received partial support from Shanghai Ninth People's Hospital, which provided the medical lab report data. We also extend our thanks to Dr. Ran Li for valuable advice and expertise in medicine.

%% file: sections/appendix.tex
\section{Appendix}
\label{sec:appendix}

\subsection{Special Medical Labels}
\label{subsec:special_medical_labels}
As is shown in the table \ref{tab:special_medical_labels} , We define a total of 13 special medical labels to convert detailed numerical values in the laboratory reports into special medical labels.

\begin{table}[h!]
    \centering
    \resizebox{0.2\textwidth}{!}{
        \begin{tabular}{l}
            \toprule
            \textbf{Labels} \\
            \midrule
            \text{[NORMAL]}        \\
            \text{[ABNORMAL]}      \\
            \text{[HI NORMAL]}      \\
            \text{[LT NORMAL]}      \\
            \text{[POSITIVE]}      \\
            \text{[NEGATIVE]}      \\
            \text{[POSITIVE]}      \\
            \text{[POSITIVE+]}      \\
            \text{[POSITIVE++]}      \\
            \text{[POSITIVE-]}      \\
            \text{[POSITIVE-]}      \\
            \text{[SENSTIVE]}      \\
            \text{[RESISTANT]}      \\
            \text{[INTERMEDIATE]}      \\
            \bottomrule
        \end{tabular}
    }
    \caption{
        \textbf{Special Medical Labels.}
    }
    \label{tab:special_medical_labels}
\end{table}

\subsection{Prompt for Zero-shot and Few-shot Generation}
\newcommand{\exbox}[2]{
{\begin{samepage}
\noindent
#1
\vspace{+0.15cm}

\nopagebreak
\noindent
\fbox{
\begin{minipage}{0.45\textwidth}{
\begin{flushleft}
\footnotesize
\texttt{\noindent #2
}
\end{flushleft}}
\end{minipage}}
\end{samepage}
}}

\exbox{Zero-shot prompt:}
{   \noindent
    Please output the patient's discharge diagnosis
    based on the given laboratory order and patient
    information. Each disease should be separated by
    a Chinese comma and then output a period. 
     Do not output anything else. Example output: 
     Low-risk mild hypertension, elevated serum uric 
     acid concentration, stage 5 chronic kidney disease. 
     Laboratory test report: \{\} Patient information: \{\}
}

\exbox{Fero-shot prompt:}
{
    Please output the patient's discharge diagnosis based on the given laboratory order and patient information. Each disease should be separated by a Chinese comma and then output a period. Finish. Do not output anything else. Examples: {} Laboratory test report: \{\} Patient information: \{\}
}

\subsection{Scalability to Input Length: Detailed Results}
\begin{figure}[h]
    \centering
    \includegraphics[width={0.5\textwidth}]{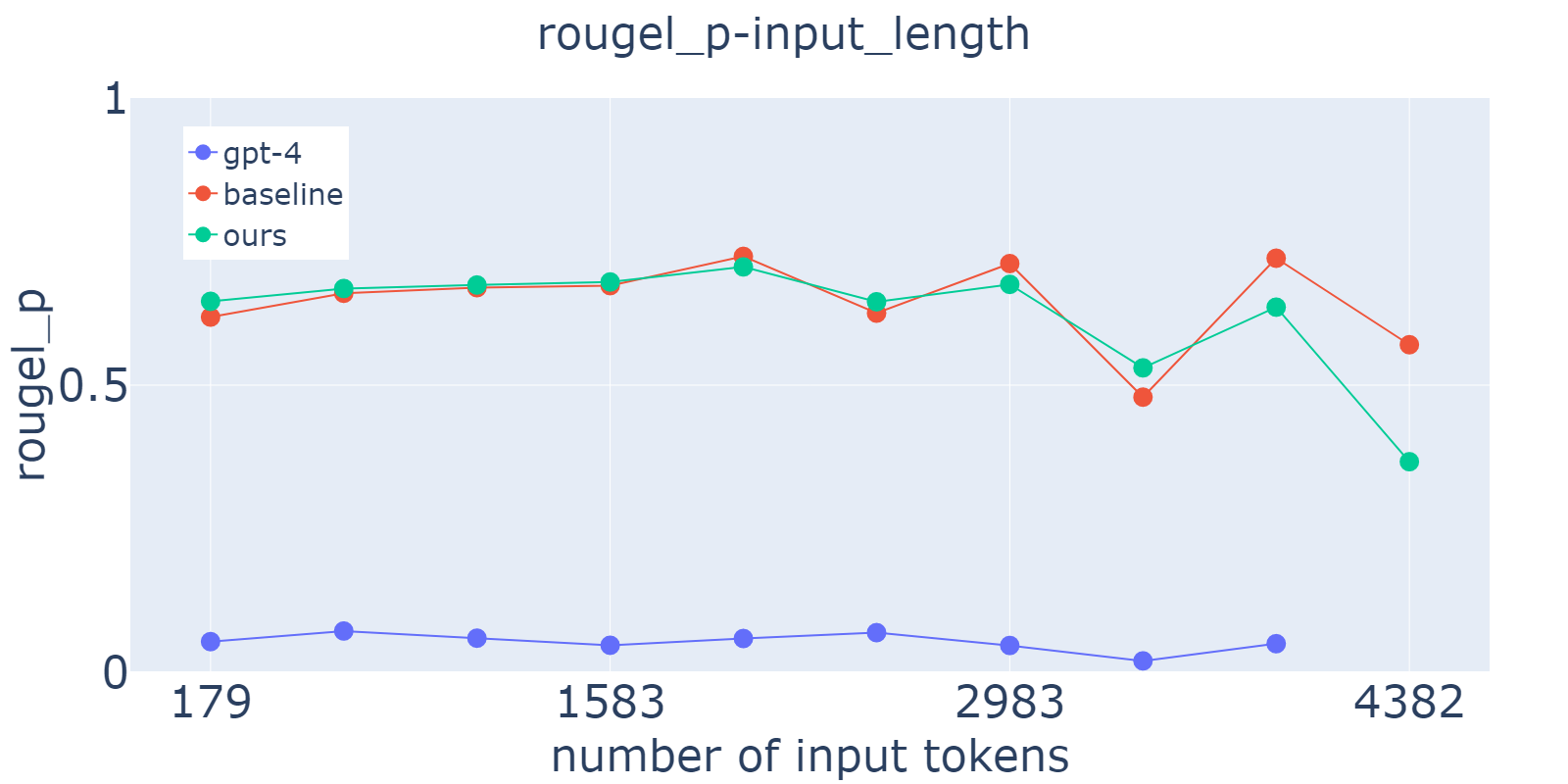}
    \caption{
        The Rouge-L precision score of different methods with respect to the number of input tokens.
    }
\end{figure}
\begin{figure}[h]
    \centering
    \includegraphics[width={0.5\textwidth}]{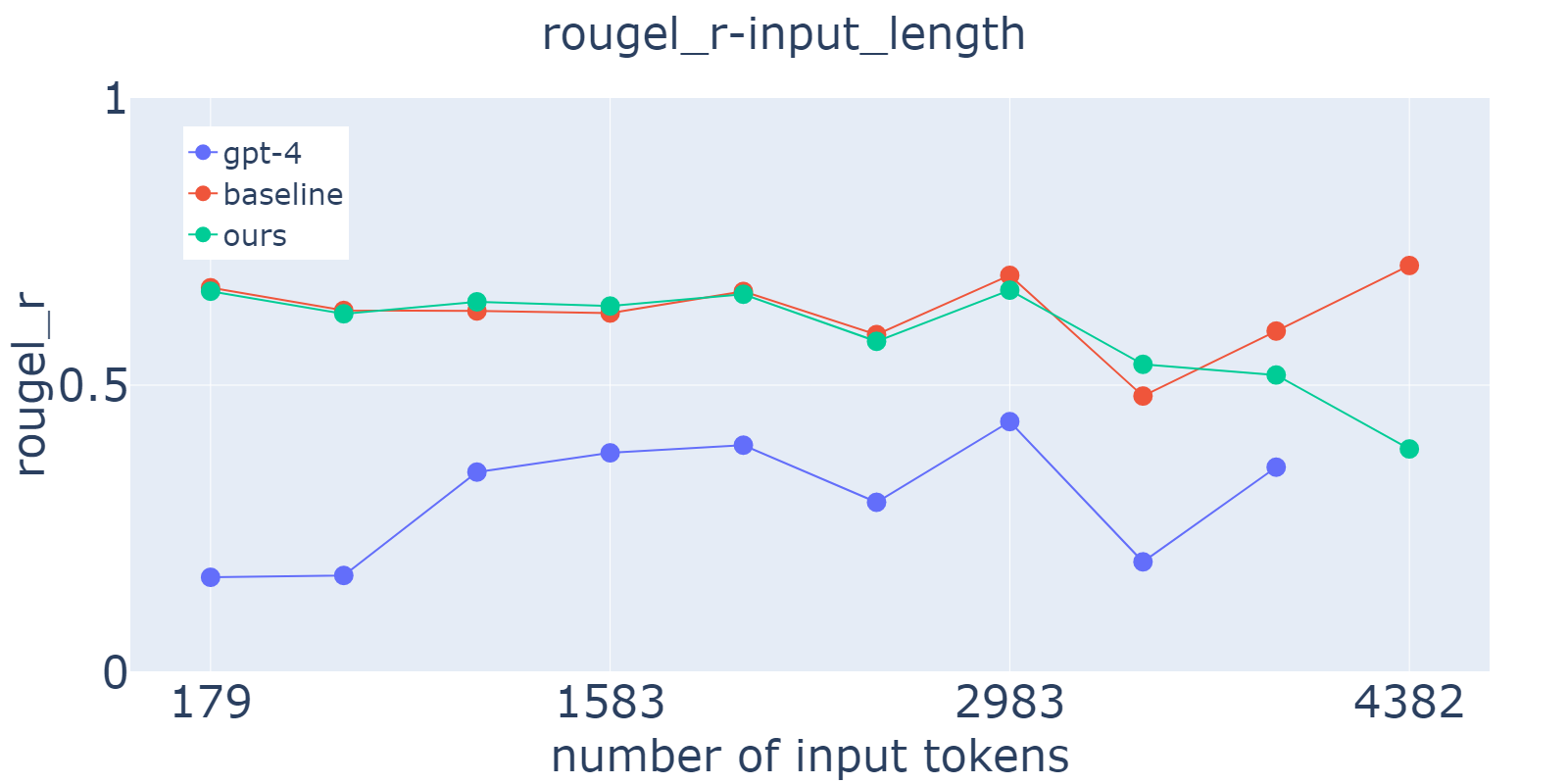}
    \caption{
        The Rouge-L recall score of different methods with respect to the number of input tokens.
    }
\end{figure}
\begin{figure}[h]
    \centering
    \includegraphics[width={0.5\textwidth}]{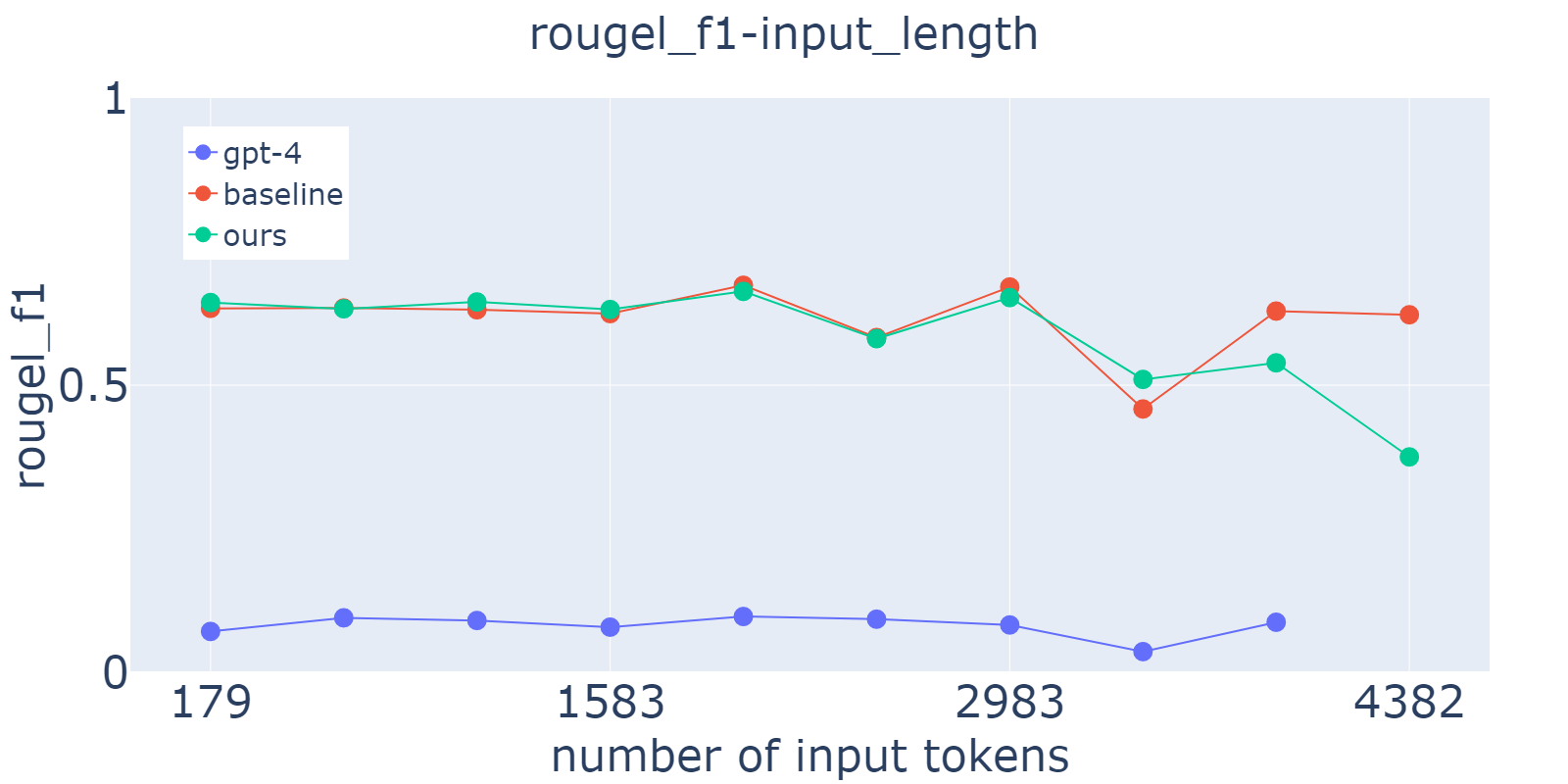}
    \caption{
        The Rouge-L F1 score of different methods with respect to the number of input tokens.
    }
\end{figure}
\begin{figure}[h]
    \centering
    \includegraphics[width={0.5\textwidth}]{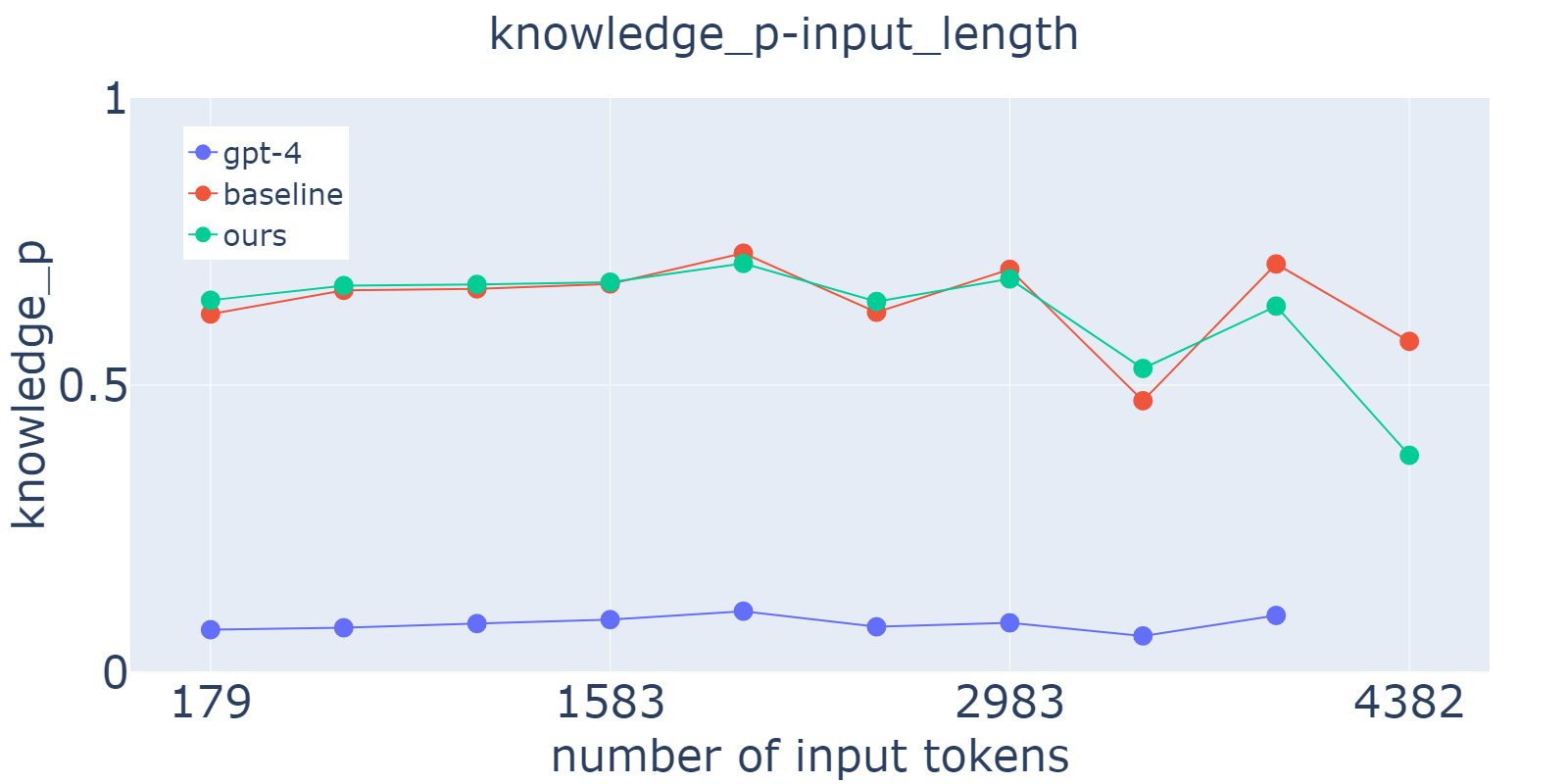}
    \caption{
        The knowledge precision score of different methods with respect to the number of input tokens.
    }
\end{figure}
\begin{figure}[h]
    \centering
    \includegraphics[width={0.5\textwidth}]{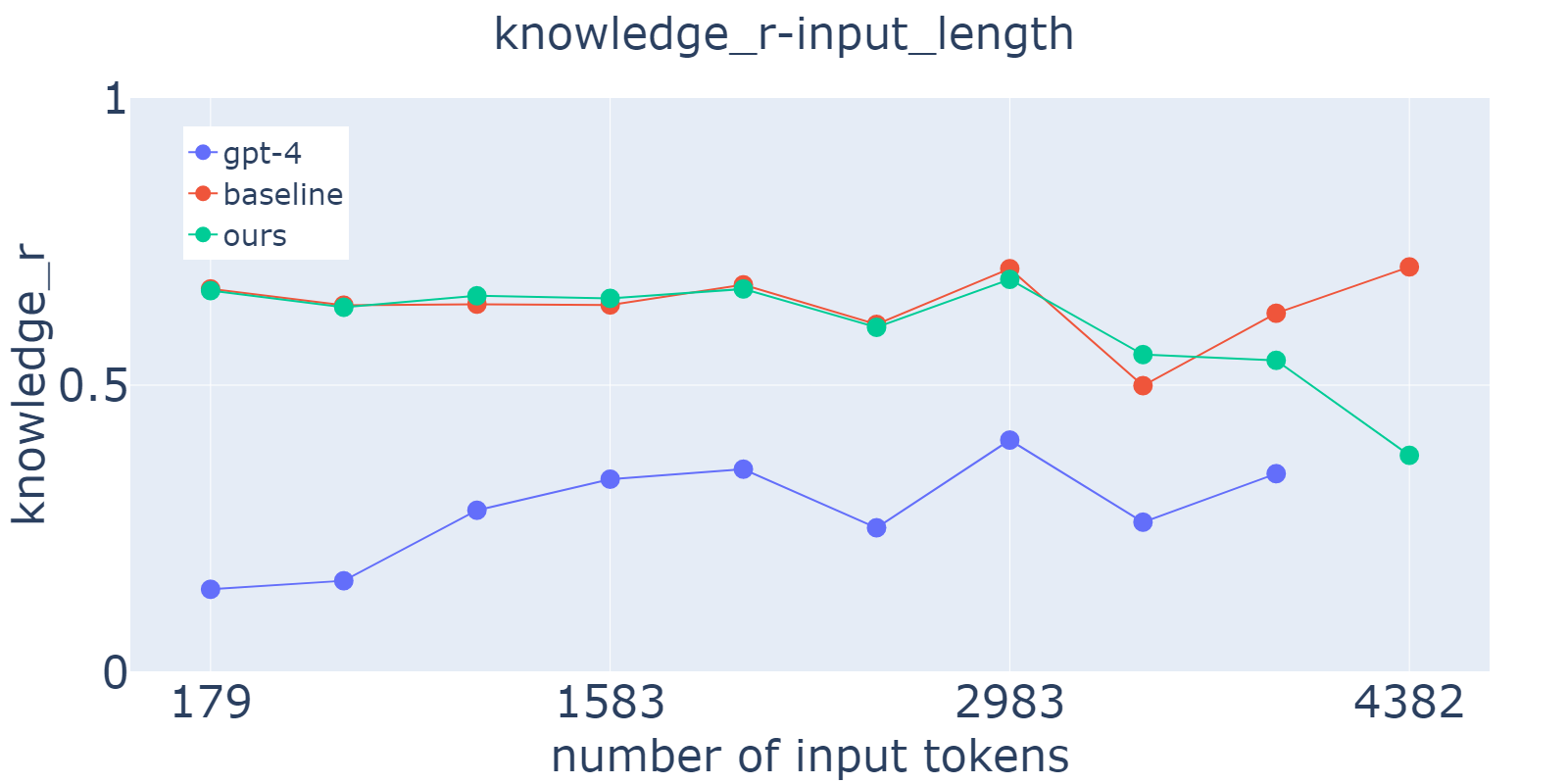}
    \caption{
        The knowledge recall score of different methods with respect to the number of input tokens.
    }
\end{figure}

%% file: main.bbl
\begin{thebibliography}{18}
\expandafter\ifx\csname natexlab\endcsname\relax\def\natexlab#1{#1}\fi

\bibitem[{KuB()}]{KuBiaoChatExcel-2023}

\newblock {{ChatExcel}}.
\newblock https://chatexcel.com/.

\bibitem[{Ope(2023)}]{OpenTabAdvancing-2023}
 2023.
\newblock {{OpenTab}}: {{Advancing Large Language Models}} as {{Open-domain Table Reasoners}}.
\newblock In \emph{The {{Twelfth International Conference}} on {{Learning Representations}}}.

\bibitem[{Chen et~al.(2023)Chen, Sarkar, Lausen, Srinivasan, Zha, Huang, and Karypis}]{HYTRELHypergraphenhanced-2023}
Pei Chen, Soumajyoti Sarkar, Leonard Lausen, Balasubramaniam Srinivasan, Sheng Zha, Ruihong Huang, and George Karypis. 2023.
\newblock \href {https://doi.org/10.48550/arXiv.2307.08623} {{{HYTREL}}: {{Hypergraph-enhanced Tabular Data Representation Learning}}}.

\bibitem[{Deng et~al.(2020)Deng, Sun, Lees, Wu, and Yu}]{TURLTable-2020}
Xiang Deng, Huan Sun, Alyssa Lees, You Wu, and Cong Yu. 2020.
\newblock \href {https://doi.org/10.48550/arXiv.2006.14806} {{{TURL}}: {{Table Understanding}} through {{Representation Learning}}}.

\bibitem[{Devlin et~al.(2018)Devlin, Chang, Lee, and Toutanova}]{BertPretraining-2018}
Jacob Devlin, Ming-Wei Chang, Kenton Lee, and Kristina Toutanova. 2018.
\newblock \href {http://arxiv.org/abs/1810.04805} {Bert: {{Pre-training}} of deep bidirectional transformers for language understanding}.
\newblock \emph{arXiv preprint arXiv:1810.04805}.

\bibitem[{Du et~al.(2022)Du, Zhang, Zhou, Wang, Zhao, Jin, Gan, Zhang, and Wipf}]{LearningEnhanced-2022}
Kounianhua Du, Weinan Zhang, Ruiwen Zhou, Yangkun Wang, Xilong Zhao, Jiarui Jin, Quan Gan, Zheng Zhang, and David Wipf. 2022.
\newblock \href {https://doi.org/10.48550/arXiv.2206.06587} {Learning {{Enhanced Representations}} for {{Tabular Data}} via {{Neighborhood Propagation}}}.

\bibitem[{Grave et~al.(2018)Grave, Joulin, and Berthet}]{UnsupervisedAlignment-2018}
Edouard Grave, Armand Joulin, and Quentin Berthet. 2018.
\newblock \href {http://arxiv.org/abs/1805.11222} {Unsupervised {{Alignment}} of {{Embeddings}} with {{Wasserstein Procrustes}}}.

\bibitem[{Han et~al.(2023)Han, Zhang, Chen, Yang, Wang, Yu, Fu, and Zhang}]{ChartLlamaMultimodal-2023}
Yucheng Han, Chi Zhang, Xin Chen, Xu~Yang, Zhibin Wang, Gang Yu, Bin Fu, and Hanwang Zhang. 2023.
\newblock \href {http://arxiv.org/abs/2311.16483} {{{ChartLlama}}: {{A Multimodal LLM}} for {{Chart Understanding}} and {{Generation}}}.

\bibitem[{Hegselmann et~al.(2023)Hegselmann, Buendia, Lang, Agrawal, Jiang, and Sontag}]{TabLLMFewshot-2023}
Stefan Hegselmann, Alejandro Buendia, Hunter Lang, Monica Agrawal, Xiaoyi Jiang, and David Sontag. 2023.
\newblock \href {https://doi.org/10.48550/arXiv.2210.10723} {{{TabLLM}}: {{Few-shot Classification}} of {{Tabular Data}} with {{Large Language Models}}}.

\bibitem[{Herzig et~al.(2020)Herzig, Nowak, M{\"u}ller, Piccinno, and Eisenschlos}]{TaPasWeakly-2020}
Jonathan Herzig, Pawel~Krzysztof Nowak, Thomas M{\"u}ller, Francesco Piccinno, and Julian Eisenschlos. 2020.
\newblock \href {https://doi.org/10.18653/v1/2020.acl-main.398} {{{TaPas}}: {{Weakly Supervised Table Parsing}} via {{Pre-training}}}.
\newblock In \emph{Proceedings of the 58th {{Annual Meeting}} of the {{Association}} for {{Computational Linguistics}}}, pages 4320--4333, {Online}. {Association for Computational Linguistics}.

\bibitem[{Lee et~al.(2019)Lee, Lee, Kim, Kosiorek, Choi, and Teh}]{SetTransformer-2019}
Juho Lee, Yoonho Lee, Jungtaek Kim, Adam~R. Kosiorek, Seungjin Choi, and Yee~Whye Teh. 2019.
\newblock \href {https://doi.org/10.48550/arXiv.1810.00825} {Set {{Transformer}}: {{A Framework}} for {{Attention-based Permutation-Invariant Neural Networks}}}.

\bibitem[{Mialon et~al.(2021)Mialon, Chen, {d'Aspremont}, and Mairal}]{TrainableOptimal-2021}
Gr{\'e}goire Mialon, Dexiong Chen, Alexandre {d'Aspremont}, and Julien Mairal. 2021.
\newblock \href {http://arxiv.org/abs/2006.12065} {A {{Trainable Optimal Transport Embedding}} for {{Feature Aggregation}} and its {{Relationship}} to {{Attention}}}.

\bibitem[{Tang et~al.(2023)Tang, Yang, Wei, Shi, Su, Cheng, Yin, and Huang}]{GraphGPTGraph-2023}
Jiabin Tang, Yuhao Yang, Wei Wei, Lei Shi, Lixin Su, Suqi Cheng, Dawei Yin, and Chao Huang. 2023.
\newblock \href {http://arxiv.org/abs/2310.13023} {{{GraphGPT}}: {{Graph Instruction Tuning}} for {{Large Language Models}}}.

\bibitem[{Touvron et~al.(2023)Touvron, Lavril, Izacard, Martinet, Lachaux, Lacroix, Rozi{\`e}re, Goyal, Hambro, Azhar, Rodriguez, Joulin, Grave, and Lample}]{LLaMAOpen-2023}
Hugo Touvron, Thibaut Lavril, Gautier Izacard, Xavier Martinet, Marie-Anne Lachaux, Timoth{\'e}e Lacroix, Baptiste Rozi{\`e}re, Naman Goyal, Eric Hambro, Faisal Azhar, Aurelien Rodriguez, Armand Joulin, Edouard Grave, and Guillaume Lample. 2023.
\newblock \href {https://doi.org/10.48550/arXiv.2302.13971} {{{LLaMA}}: {{Open}} and {{Efficient Foundation Language Models}}}.

\bibitem[{Yang et~al.(2022)Yang, Gupta, Upadhyay, He, Goel, and Paul}]{TableFormerRobust-2022}
Jingfeng Yang, Aditya Gupta, Shyam Upadhyay, Luheng He, Rahul Goel, and Shachi Paul. 2022.
\newblock \href {http://arxiv.org/abs/2203.00274} {{{TableFormer}}: {{Robust Transformer Modeling}} for {{Table-Text Encoding}}}.

\bibitem[{Ye et~al.(2023)Ye, Lu, Wang, Li, Wu, Chen, and Zhao}]{CTBERTLearning-2023}
Chao Ye, Guoshan Lu, Haobo Wang, Liyao Li, Sai Wu, Gang Chen, and Junbo Zhao. 2023.
\newblock \href {https://doi.org/10.48550/arXiv.2307.04308} {{{CT-BERT}}: {{Learning Better Tabular Representations Through Cross-Table Pre-training}}}.

\bibitem[{Zeng et~al.(2022)Zeng, Liu, Du, Wang, Lai, Ding, Yang, Xu, Zheng, Xia, Tam, Ma, Xue, Zhai, Chen, Zhang, Dong, and Tang}]{GLM130BOpen-2022}
Aohan Zeng, Xiao Liu, Zhengxiao Du, Zihan Wang, Hanyu Lai, Ming Ding, Zhuoyi Yang, Yifan Xu, Wendi Zheng, Xiao Xia, Weng~Lam Tam, Zixuan Ma, Yufei Xue, Jidong Zhai, Wenguang Chen, Peng Zhang, Yuxiao Dong, and Jie Tang. 2022.
\newblock \href {https://doi.org/10.48550/arXiv.2210.02414} {{{GLM-130B}}: {{An Open Bilingual Pre-trained Model}}}.

\bibitem[{Zhu et~al.(2024)Zhu, Liu, Feng, Wang, Li, and Chua}]{TATLLMSpecialized-2024}
Fengbin Zhu, Ziyang Liu, Fuli Feng, Chao Wang, Moxin Li, and Tat-Seng Chua. 2024.
\newblock \href {http://arxiv.org/abs/2401.13223} {{{TAT-LLM}}: {{A Specialized Language Model}} for {{Discrete Reasoning}} over {{Tabular}} and {{Textual Data}}}.

\end{thebibliography}
